\documentclass[conference]{IEEEtran}
\IEEEoverridecommandlockouts

\usepackage{cite}
\usepackage[T1]{fontenc}
\usepackage[utf8]{inputenc}
\usepackage{amsmath,amssymb,amsfonts}
\usepackage{algorithmic}
\usepackage{graphicx}
\usepackage{textcomp}
\usepackage{xcolor}
\usepackage{verbatim}

\usepackage{comment}
\usepackage[table]{xcolor}
\usepackage{booktabs}
\usepackage{multirow}
\def\BibTeX{{\rm B\kern-.05em{\sc i\kern-.025em b}\kern-.08em
    T\kern-.1667em\lower.7ex\hbox{E}\kern-.125emX}}
\begin{document}

\title{Reference-Based Prosody and Rhythm Evaluation for Spoken Dialogue Systems\\
\thanks{This work was supported by the SPEAR project. Add full funding acknowledgments before submission.}
}

\author{\IEEEauthorblockN{Ashish Hallur\IEEEauthorrefmark{1}, Thomas Thebaud\IEEEauthorrefmark{1}, Georgi Tinchev\IEEEauthorrefmark{2}, Venkatesh Ravichandran\IEEEauthorrefmark{2}, Laureano Moro-Velazquez\IEEEauthorrefmark{1}}
\IEEEauthorblockA{\IEEEauthorrefmark{1}\textit{Department of Electrical and Computer Engineering, Johns Hopkins University}, Baltimore, MD, USA\\
Email: ahallur1@jhu.edu}
\IEEEauthorblockA{\IEEEauthorrefmark{2}\textit{Amazon Inc.}, Seattle, WA, USA}}

\maketitle

\begin{abstract}
Speech-to-speech (S2S) AI agents are advancing rapidly, yet evaluation lacks interpretable speech-native measures for conversational prosody and rhythm. 
Because $F_0$, speaking rate, articulation rate, and pausing shift with model-predicted speaker traits and interaction state, pooled human statistics can be poorly calibrated for evaluating a particular output.
Using 4000+ hours of dyadic English conversation from the Seamless Interaction dataset, we construct matched reference regimes for $F_0$ mean, $F_0$ expressivity, speech rate, articulation rate, pause ratio, and mean pause duration.
We then define a percentile-based evaluation protocol: extract the same metrics from an S2S output waveform, compare them to the closest matched human reference stratum, and report percentile deviations or 5th--95th percentile out-of-regime flags.
On held-out human rows, pooled references over-flag state-conditioned $F_0$ expressivity and rhythm, while matched references return flag rates closer to the nominal 10\% and make deviation direction interpretable.
These outputs serve as behavioral plausibility checks that complement, rather than replace, perceptual and user-centered evaluation.
\end{abstract}


\section{Introduction}

Spoken conversation is a highly coordinated form of communication, where interlocutors manage turn-taking while continuously adapting prosody and timing to the unfolding interaction \cite{riest_anticipation_2015, levinson_turn-taking_2016}. 
Despite this complexity, transitions are typically smooth and supported by systematic cues, including prosodic and syntactic structure \cite{koiso_analysis_1998}.
Speech technology has progressed toward S2S AI agents capable of producing fluent spoken responses. 
Evaluation, however, is still dominated by text-based measures, task success, or subjective ratings, which do not directly quantify speech-native interactional behavior. 
Systems trained primarily on read speech often fail to reproduce properties characteristic of spontaneous dialogue \cite{omahony_combining_2022}.

Conversational speech differs from read speech across multiple dimensions, including prosodic range and stress realization \cite{howell1991comparison, hazan2010does}. 
Temporal organization varies with context and listener demands, shaping speaking rate and rhythmic structure \cite{tauroza_speech_1990, ward_automatic_2004, dowding_user_2024, xie_speaking_2024}. 
These patterns align with accounts of dialogue as adaptive coordination, where speakers modify behavior to meet various contextual demands \cite{fusaroli_dialog_2014, wynn_conversational_2024}.

Prosodic variation, particularly fundamental frequency ($F_0$), encodes communicative intent, emotion, and discourse structure, and interacts with prominence and stress cues \cite{banziger_role_2005, pell_emotional_2011, xu_speech_2005, sluijter_acoustic_1996, sluijter_spectral_1996}. 
Temporal structure is central to conversational timing, with systematic patterns in pausing and speaking rate \cite{goldman1958speech, pellegrino_automatic_2004, pellegrino_across-language_2011, jiao_estimating_2015, morgan_combining_1998, nanjo_language_2004}. 
Recent work also shows that conversational speech behaviors are context dependent, motivating analyses that explicitly condition on interaction factors \cite{wynn_conversational_2024}.

Prior work has often examined prosody and timing in isolation or at a limited scale, leaving a gap in large-scale, multidimensional reference regimes for conversational speech technology. 
This gap is limiting for spoken dialogue evaluation, where a system may fall inside a pooled range while still producing prosody or timing that is implausible for the relevant speaker profile or interactional state.

We frame large-scale conversational distributions as an operational evaluation resource rather than as descriptive baselines alone. 
Using robust percentile-based $F_0$ measures and rhythm metrics from over 4{,}000 hours of English dyadic conversation, we make four contributions. 
First, we construct matched reference regimes for conversational $F_0$ mean, $F_0$ expressivity, speech rate, articulation rate, pause ratio, and mean pause duration. 
Second, we show that pooled references can obscure systematic shifts associated with model-predicted sex label, model-predicted age bin, arousal, and dominance \cite{wynn_conversational_2024, traunmuller_acoustic_2000}. 
Third, we define a simple evaluation procedure that extracts the same metrics from an S2S output waveform and reports percentile deviations or out-of-regime flags relative to the closest matched human reference stratum. 
Fourth, we provide reproducible reference tables and an extraction/comparison protocol for speech-native behavioral plausibility checks in spoken dialogue systems.

\section{Methods}

\subsection{Dataset}
We analyze dyadic conversational speech from the Seamless Interaction dataset, a large-scale corpus of face-to-face interactions designed to capture both Naturalistic interactions with untrained participants and Improvised interactions with trained actors. 
The released dataset contains 4,065.04 hours of interaction time, comprising 64,739 interactions segmented from 5,098 one-hour recording sessions involving 4,284 participants \cite{seamless_interaction}. 
We use ``interaction'' to refer to a single conversational segment within a session. 
Each interaction yields two speaker channels, and all metrics are computed at the channel level. 
The dataset includes Naturalistic interactions with untrained participants and Improvised interactions with trained actors, supporting analyses that condition on interaction factors \cite{fusaroli_dialog_2014, wynn_conversational_2024}.

\subsection{Prosodic Metrics}
Fundamental frequency ($F_0$) is defined only during voiced speech, so each audio signal is analyzed frame-by-frame using the autocorrelation-based $F_0$ estimator in Praat via parselmouth \cite{jadoul2018introducing}. 
$F_0$ extraction uses conservative bounds (75--500 Hz) to cover typical adult conversational ranges while reducing octave errors and spurious tracks \cite{vogel2009standardization}. 
We define the voiced ratio as the proportion of frames assigned a nonzero $F_0$ value and exclude speaker-channels with a voiced ratio $<0.05$ \cite{titze_toward_1994}. 
To reduce sensitivity to spontaneous-speech artifacts and $F_0$-tracking outliers, we compute percentile-trimmed summaries by retaining voiced $F_0$ values between the 10th and 90th percentiles within each speaker-channel \cite{nguyen2024culturax}. 
We report the 10--90\% trimmed mean, standard deviation, and range of $F_0$.

\subsection{Temporal Metrics}\label{sec:temporalmetrics}
Temporal metrics are computed from word-level timestamps in the dataset's ASR-aligned transcripts and the Voice Activity Detection (VAD) segments distributed with the corpus \cite{seamless_interaction}. 
To obtain stable long-term estimates of rate, we follow evidence that speaking-rate estimates stabilize over an average stabilization time of 12.1 s (most values between 7.9 and 16.2 s) \cite{arantes2018minimum}. 
Accordingly, we define speech-activity stretches using the provided VAD, merge adjacent segments separated by at most 1.0 s, and retain only continuous stretches with duration $\geq 12.1$ s before computing speaker-channel-level temporal statistics. 
We define pauses as inter-word gaps $\geq 0.2$ s, a common practical threshold because shorter silences are difficult to distinguish from stop closures and their inclusion increases annotation and measurement burden \cite{campione2002large}. 
Let $W$ be the number of retained words, $T$ the total duration (sum of retained stretch durations), and $P$ the total pause time (sum of inter-word gaps $\geq 0.2$ s within stretches). 
In semi-spontaneous speech, WPM showed a very strong correlation with naïve listeners' tempo ratings \cite{iwarsson2023measuring}, and hence we report speaking rates in words per minute (WPM) because it closely tracks perceived speech tempo and is directly measurable from word-aligned timestamps in the dataset.
We report speech rate $=60\cdot \frac{W}{T}$ and articulation rate $=60\cdot \frac{W}{(T-P)}$ \cite{goldman1956determinants} in WPM, along with pause ratio $=\frac{P}{T}$.

\begin{table}[t]
\centering
\caption{Pooled operating regimes (speaker-channel level). For each track, $N$ and interaction-hours are constant across metrics and are reported in the header row.}
\label{tab:pooled_regimes_single}
\small
\setlength{\tabcolsep}{4pt}
\begin{tabular}{l r l r}
\toprule
\textbf{Metric} & \textbf{Median} & \textbf{IQR (25--75\%)} & \textbf{Mean} \\
\midrule
\rowcolor{gray!20}\multicolumn{4}{c}{ \textbf{Prosody} (\,$N{=}121{,}813$,\, $3{,}863$h)} \\
$F_0$ Mean [Hz]       & 157.4  & 120.1--198.6 & 161.5 \\
$F_0$ SD [Hz] & 20.84  & 13.79--30.07 & 23.22 \\
$F_0$ Range [Hz]       & 87.11  & 57.11--125.8 & 95.82 \\
\midrule
\rowcolor{gray!20}\multicolumn{4}{c}{\textbf{Temporal} (\,$N{=}91{,}471$,\, $3{,}045$h)} \\
Speech rate [wpm]        & 175.9  & 156.0--195.9 & 175.8 \\
Articulation rate [wpm]  & 237.8  & 216.1--259.5 & 237.2 \\
Pause ratio              & 0.2575 & 0.2166--0.2996 & 0.2595 \\
Mean pause duration [s]  & 0.5845 & 0.5225--0.6559 & 0.6058 \\
\bottomrule
\end{tabular}
\end{table}

\subsection{Speaker Trait Annotations (Vox-Profile)}
\label{sec:voxprofile}
Since Seamless Interaction does not include ground-truth speaker-trait metadata, we augment Seamless Interactions' metadata with model-conditioned speaker-trait and interaction-state annotations using Vox-Profile, a benchmark and toolchain for characterizing static and dynamic speech traits using speech foundation models \cite{feng2025vox}.
For each speaker-channel waveform, we first resample the audio to 16 kHz and apply Silero VAD \cite{Silero_VAD} to extract speech-only material. 
We then run two WavLM-based predictors released with Vox-Profile -- a multitask age/sex model and a dimensional emotion model. 
The age/sex model outputs an age estimate (mapped to model-predicted age bins for analysis) and a binary sex prediction with an associated posterior probability, as the original dataset does not provide these speaker traits, while the emotion model outputs continuous arousal, valence, and dominance scores in $[0,1]$.
Vox-Profile reports high performance for sex classification (97.7\% acc., macro-F1 0.971) and moderate performance for age-bin prediction (67.6\% acc., macro-F1 0.624) \cite{feng2025vox}.

We focus on model-predicted sex label, model-predicted age bin, arousal, and dominance as stratification variables because they capture complementary, operationally relevant sources of variation for conversational prosody and rhythm. 
Model-predicted sex label is expected to strongly track habitual $F_0$-related statistics associated with anatomical and physiological voice differences \cite{traunmuller_acoustic_2000, titze_toward_1994}. 
Chronological age has been linked to systematic differences in conversational timing and fluency-related measures such as speaking rate and pausing, motivating the use of model-predicted age bin as an operational stratifier \cite{tauroza_speech_1990, goldman1958speech, goldman1961significance}. 
Arousal and dominance provide continuous proxies for interactional state that modulate prosodic expressivity and temporal pacing in natural speech \cite{banziger_role_2005, pell_emotional_2011, banse1996acoustic}. 
In preliminary screening across available model-conditioned speaker traits and interaction-state variables, these four factors produced the most consistent and interpretable shifts in the prosodic and temporal distributions studied here, and we therefore center them in a compact, evaluation-oriented analysis.

\subsection{Pooled vs. Matched Evaluation Check}
To test whether matched regimes improve evaluation calibration, we split speaker-channel rows into deterministic participant-held-out calibration and evaluation sets. 
For each metric, we estimate 5th--95th percentile thresholds from the calibration split under two references: a pooled reference using all usable calibration rows, and a matched reference using the relevant stratum. 
We match mean $F_0$ by model-predicted sex label and match $F_0$ expressivity and temporal metrics by arousal or dominance sextile. 
We then report the held-out evaluation percentage falling outside each interval; a calibrated 5th--95th percentile reference should flag approximately 10\% of human conversational rows.

\section{Results}\label{sec:regimes}

\begin{table}[t]
\centering
\caption{Model-predicted sex-label effects across prosodic and temporal metrics. Values are computed on the usable prosody subset (Fig.~1; $N{=}121{,}813$) and temporal subset (Fig.~4; $N{=}91{,}471$, $3{,}045$h). Effect size is Cliff's $\delta$ (Male vs. Female; Mann--Whitney $U$).}
\label{tab:sex_label_effects}
\small
\setlength{\tabcolsep}{3pt}
\begin{tabular}{l l c c r}
\toprule
\textbf{Metric} & \textbf{Group} & \multicolumn{2}{c}{\textbf{Values}} \\
(\textbf{Cliff's $\delta$}) & \textbf{Label} & \textbf{Median} & \textbf{Mean} & \textbf{Unit}\\
\midrule
\rowcolor{gray!20}\multicolumn{5}{c}{\textbf{Prosody (Fig.~1)}} \\
Mean $F_0$ (10--90\%)     & Male   & 121.9 & 125.7 & Hz\\
($\delta=-0.957$)         & Female & 200.7 & 202.7 & Hz\\
$F_0$ SD (10--90\%)       & Male   & 15.00 & 17.82 & Hz\\
($\delta=-0.635$)         & Female & 27.81 & 29.44 & Hz\\
$F_0$ range (10--90\%)    & Male   & 61.95 & 73.44 & Hz\\
($\delta=-0.644$)         & Female & 116.8 & 121.6 & Hz\\
\midrule
\rowcolor{gray!20}\multicolumn{5}{c}{\textbf{Temporal (Fig.~4)}} \\
Speech rate              & Male   & 177.64 & 177.50 & wpm\\
($\delta=0.066$)         & Female & 173.95 & 173.94 & wpm\\
Articulation rate        & Male   & 242.94 & 242.02 & wpm\\
($\delta=0.180$)         & Female & 232.55 & 231.73 & wpm\\
Pause ratio              & Male   & 0.2656 & 0.2673 & --\\
($\delta=0.155$)         & Female & 0.2483 & 0.2505 & --\\
Mean pause duration      & Male   & 0.594 & 0.616 & s\\
($\delta=0.118$)         & Female & 0.573 & 0.594 & s\\
\bottomrule
\end{tabular}
\end{table}

\begin{table}[t]
\centering
\caption{State and model-predicted age-bin effects on prosodic expressivity and conversational timing. Effects are Spearman $\rho$ for arousal/dominance and Kruskal--Wallis $\epsilon^2$ for model-predicted age bins. All reported effects are statistically significant (two-sided; $p\approx 0$ at this scale). Usable subsets match Figs.~2--4 (prosody/state: $N{=}121{,}813$; temporal/state: $N{=}91{,}471$) and Fig.~5 (age-bin: prosody $N{=}118{,}033$; temporal $N{=}88{,}287$).}
\label{tab:state_age_effects}
\small
\setlength{\tabcolsep}{4pt}
\begin{tabular}{l c}
\toprule
\textbf{Metric} & \textbf{Effect} \\
\midrule
\rowcolor{gray!20}\multicolumn{2}{c}{\textbf{Arousal effects (Spearman $\rho$)}} \\
$F_0$ SD (10--90\%)            & $\rho=0.544$ \\
$F_0$ range (10--90\%)         & $\rho=0.516$ \\
Speech rate [wpm]              & $\rho=0.187$ \\
Pause ratio                    & $\rho=-0.170$ \\
\midrule
\rowcolor{gray!20}\multicolumn{2}{c}{\textbf{Dominance effects (Spearman $\rho$)}} \\
$F_0$ SD (10--90\%)            & $\rho=0.463$ \\
$F_0$ range (10--90\%)         & $\rho=0.430$ \\
Speech rate [wpm]              & $\rho=0.200$ \\
Pause ratio                    & $\rho=-0.194$ \\
\midrule
\rowcolor{gray!20}\multicolumn{2}{c}{\textbf{Model-predicted age-bin effects (Kruskal--Wallis $\epsilon^2$)}} \\
Mean $F_0$ (10--90\%)          & $\epsilon^2=0.00643$ \\
Speech rate [wpm]              & $\epsilon^2=0.01623$ \\
Pause ratio                    & $\epsilon^2=0.00511$ \\
\bottomrule
\end{tabular}
\end{table}

\begin{table}[t]
\centering
\caption{Naturalistic vs. Improvised subset comparison. Medians are in Hz for $F_0$, wpm for rates, seconds for mean pause duration, and unitless for pause ratio. Effect size is Cliff's $\delta$ (Naturalistic vs. Improvised).}
\label{tab:naturalistic_improvised}
\small
\setlength{\tabcolsep}{3pt}
\begin{tabular}{l r r r l}
\toprule
\textbf{Metric} & \textbf{Nat.} & \textbf{Imp.} & \textbf{$\delta$} & \textbf{Interp.} \\
\midrule
$F_0$ mean        & 149.18 & 173.80 & -0.198 & small \\
$F_0$ SD          & 19.07  & 25.51  & -0.278 & small \\
$F_0$ range       & 80.07  & 104.34 & -0.254 & small \\
Speech rate       & 176.45 & 174.51 & 0.031  & negligible \\
Articulation rate & 239.57 & 234.28 & 0.088  & negligible \\
Pause ratio       & 0.2596 & 0.2525 & 0.070  & negligible \\
Mean pause dur.   & 0.592  & 0.568  & 0.136  & negligible \\
\bottomrule
\end{tabular}
\end{table}

\begin{table}[t]
\centering
\caption{Held-out out-of-regime flag rates under pooled and matched 5th--95th percentile references. A calibrated reference should flag about 10\% of human rows. Matched references use model-predicted sex label for mean $F_0$ and state sextiles for expressivity/rhythm.}
\label{tab:pooled_matched_flags}
\small
\setlength{\tabcolsep}{3pt}
\begin{tabular}{l r r r}
\toprule
\textbf{Evaluation group} & \textbf{Eval. $N$} & \textbf{Pooled} & \textbf{Matched} \\
 & & \textbf{flag \%} & \textbf{flag \%} \\
\midrule
Male mean $F_0$         & 32{,}373 & 9.88  & 12.06 \\
Female mean $F_0$       & 28{,}047 & 8.63  & 8.35 \\
Low-arousal $F_0$ SD    & 10{,}672 & 21.11 & 10.16 \\
High-arousal $F_0$ SD   & 10{,}697 & 16.07 & 9.54 \\
Low-dominance $F_0$ SD  & 10{,}863 & 18.30 & 8.87 \\
High-dominance $F_0$ SD & 10{,}630 & 15.41 & 9.72 \\
High-arousal speech rate & 7{,}177  & 15.76 & 12.50 \\
High-arousal pause ratio & 7{,}177  & 13.18 & 10.84 \\
\bottomrule
\end{tabular}
\end{table}

\begin{figure*}[t]
  \centering
  \includegraphics[width=\textwidth]{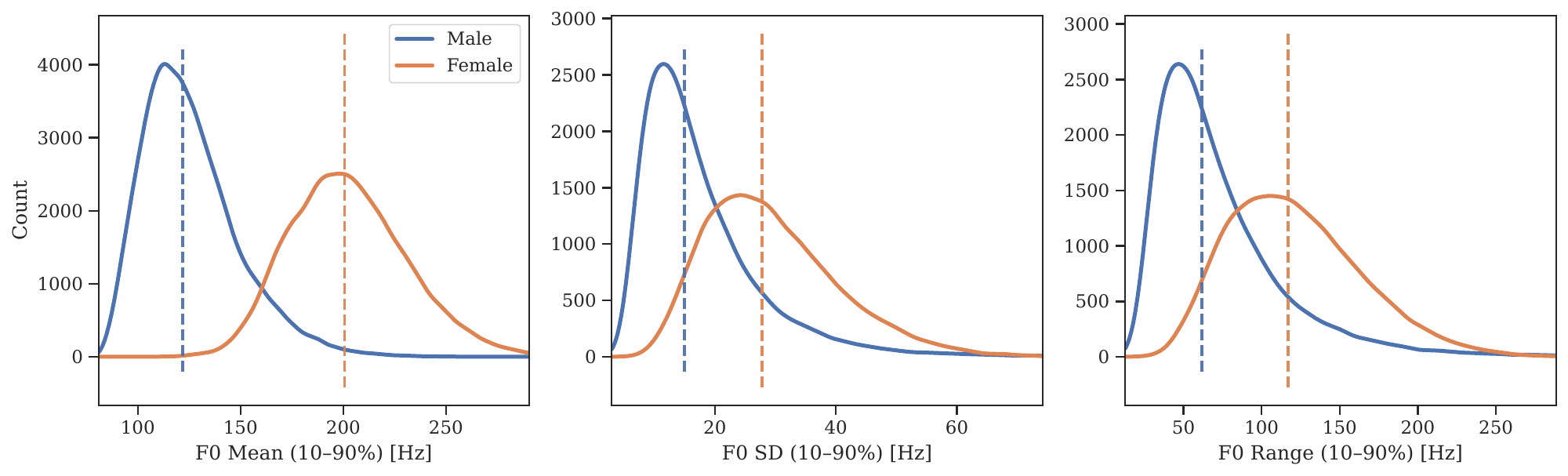}
  \caption{Prosodic operating regimes (10--90\% trimmed) stratified by model-predicted sex label. Kernel density curves (scaled to counts: $N{=}121{,}813$; Male $=65{,}214$, Female $=56{,}599$) are shown for $F_0$ Mean, SD, and range. Vertical dashed lines mark group medians.}
  \label{fig:prosody_sex_label}
\end{figure*}

\begin{figure}[t]
  \centering
  \includegraphics[width=\linewidth]{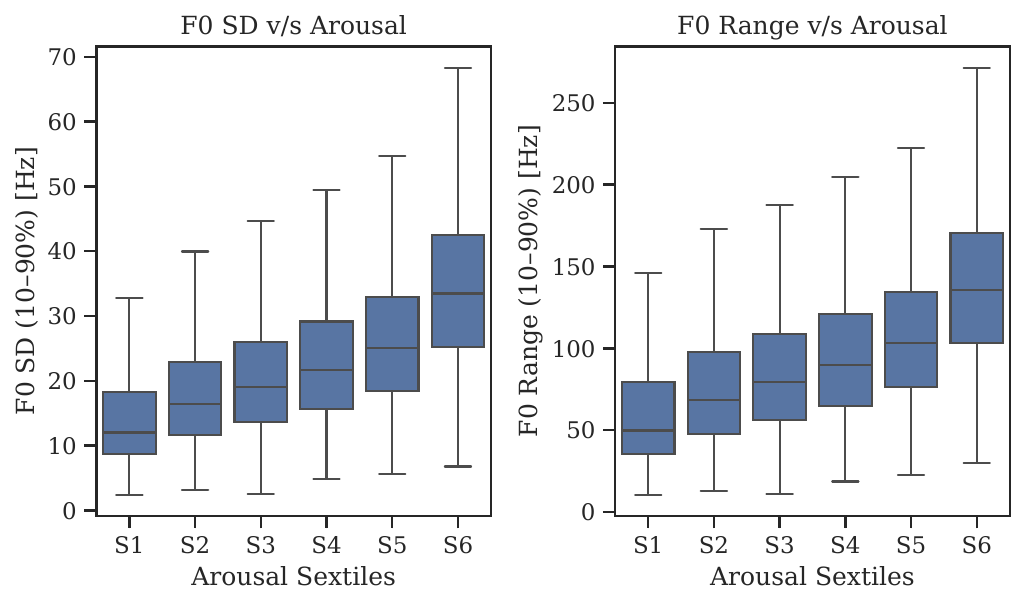}
  \caption{State-driven $F_0$ expressivity across arousal sextiles (S1--S6). Boxplots summarize 10--90\% trimmed $F_0$ standard deviation and $F_0$ range within arousal bins ($N{=}121{,}813$).}
  \label{fig:prosody_arousal}
\end{figure}

\begin{figure}[t]
  \centering
  \includegraphics[width=\linewidth]{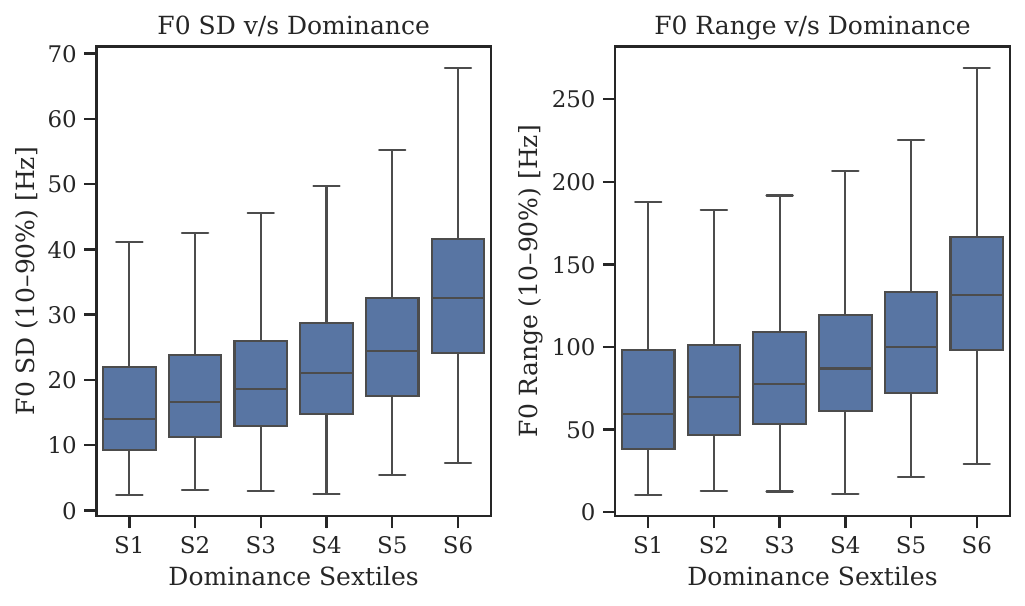}
  \caption{$F_0$ expressivity across dominance sextiles (S1--S6). Boxplots summarize 10--90\% trimmed $F_0$ standard deviation and $F_0$ range within dominance bins ($N{=}121{,}813$).}
  \label{fig:prosody_dominance}
\end{figure}

We summarize the operating regimes of conversational prosody and rhythm and quantify how these regimes vary with speaker and interaction factors.
Throughout, $N$ denotes the number of usable speaker-channel samples satisfying the subset criteria for a given figure/table, and hours denote total speech duration aggregated over those usable samples (reported as interaction-time). 
We use the term ``expressivity'' to jointly refer to the range and standard deviation of $F_0$.
Table~\ref{tab:pooled_regimes_single} reports pooled reference ranges, Table~\ref{tab:sex_label_effects} summarizes two-group model-predicted sex-label effects, Table~\ref{tab:state_age_effects} summarizes continuous state (arousal, dominance) and model-predicted age-bin effects, and Table~\ref{tab:naturalistic_improvised} checks whether the Naturalistic and Improvised portions of Seamless Interaction require separate treatment. 
Table~\ref{tab:pooled_matched_flags} gives the evaluation consequence of these shifts: pooled references substantially over-flag low- and high-state $F_0$ expressivity, while matched references return held-out human rows close to the nominal 10\% flag rate. 
Since the corpus is large, we report effect sizes in addition to significance tests: for two-group comparisons we use the Mann--Whitney $U$ test with Cliff's $\delta$ to quantify distributional separation without assuming normality \cite{cliff1993dominance}; for continuous co-variates (arousal, dominance) we use Spearman rank correlation $\rho$ to capture monotonic associations robustly \cite{spearman1961proof}; and for age-bin comparisons we use the Kruskal--Wallis test with $\epsilon^2$ as an effect size for multi-group differences \cite{kruskal1952use, tomczak2014need}. 
The subset comparison shows negligible temporal differences and only small $F_0$ differences (maximum $|\delta|=0.278$), which are much weaker than the dominant speaker- and state-conditioned shifts; we therefore pool these subsets for the main reference characterization.

\subsection{Prosodic Regimes}
We characterize conversational prosody using robust 10--90\% trimmed $F_0$ statistics that summarize $F_0$ mean, SD, and range.
Figure~\ref{fig:prosody_sex_label} shows that the mean $F_0$ is strongly conditioned by model-predicted sex label, with large distributional separation that makes pooled absolute-$F_0$ targets inappropriate for evaluation. 
The held-out check in Table~\ref{tab:pooled_matched_flags} shows that pooled mean-$F_0$ flags are also directionally biased: male rows are flagged almost entirely below the pooled interval (9.8\% low vs. 0.1\% high), while female rows are flagged above it (0.0\% low vs. 8.6\% high). 
This aligns with long-standing evidence that anatomical/physiological differences yield distinct habitual $F_0$ regimes across speakers \cite{traunmuller_acoustic_2000, titze_toward_1994}.

Beyond this baseline conditioning, Figures~\ref{fig:prosody_arousal} and \ref{fig:prosody_dominance} show a consistent scaling of prosodic expressivity with interactional state. 
In particular, $F_0$ SD and $F_0$ range increase monotonically across arousal and dominance sextiles obtained with VoxProfile. 
This matches prior work showing that natural speech tends to use a wider $F_0$ bandwidth in higher-activation emotional or interactional states, with larger pitch excursions when speakers are more activated or more socially assertive \cite{banziger_role_2005, banse1996acoustic, geng2020acoustic}. 
If the $F_0$ produced by a speaker or an S2S dialogue system stays in a narrow-band variation in high-arousal or high-dominance contexts, it may sound constrained or unnatural even when the mean $F_0$ is within a typical range \cite{liu2021multiple}.
Prosodic evaluation should therefore include expressivity-sensitive checks in addition to $F_0$ level, and interpret deviations relative to reference distributions stratified by arousal and dominance \cite{pell_emotional_2011, xu_speech_2005}. 
In held-out rows, a pooled $F_0$-SD reference flags 21.11\% of low-arousal and 16.07\% of high-arousal samples, whereas arousal-matched references reduce these rates to 10.16\% and 9.54\%, respectively.

\subsection{Temporal Regimes}

\begin{figure*}[t]
  \centering
  \includegraphics[width=\textwidth]{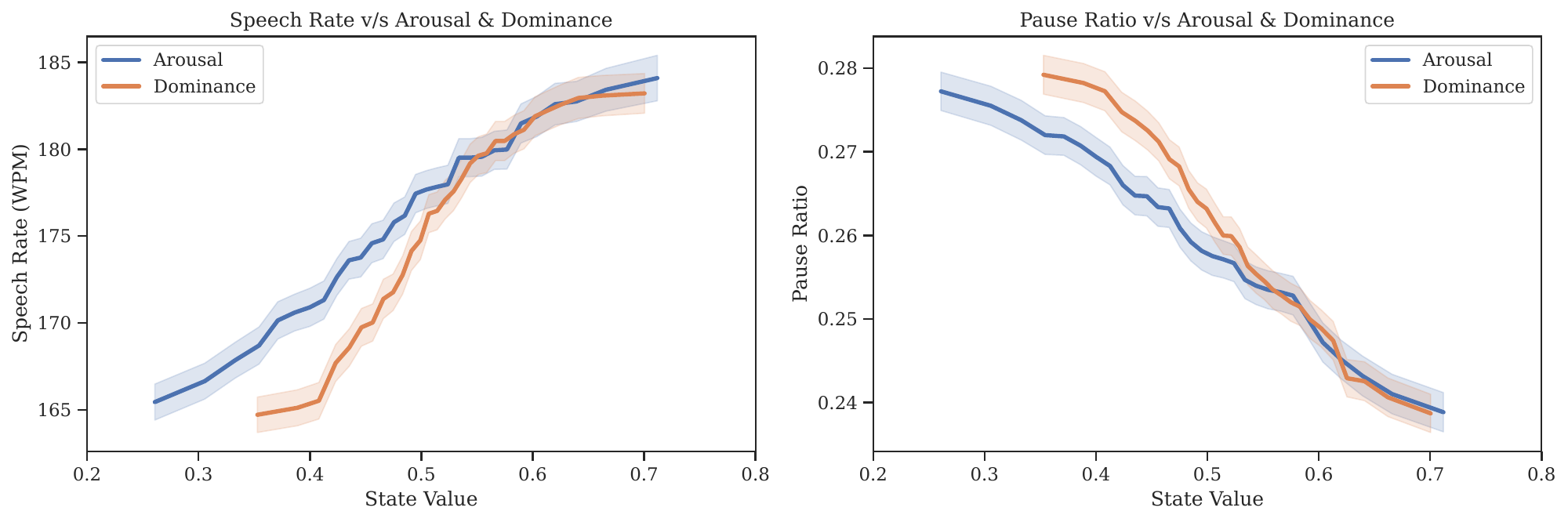}
  \caption{Conversational rhythm varies with interactional state. Smoothed trends show speech rate (left) and pause ratio (right) as functions of arousal and dominance (0--1). Trends are computed from equal-count bins ($N{=}91{,}471$) over the state axis, followed by smoothing. Shaded bands indicate $\approx95\%$ uncertainty (SEM-based). Bins with sparse data are omitted to reduce unstable end effects.}
  \label{fig:temporal_state}
\end{figure*}

\begin{figure*}[t]
  \centering
  \includegraphics[width=\textwidth]{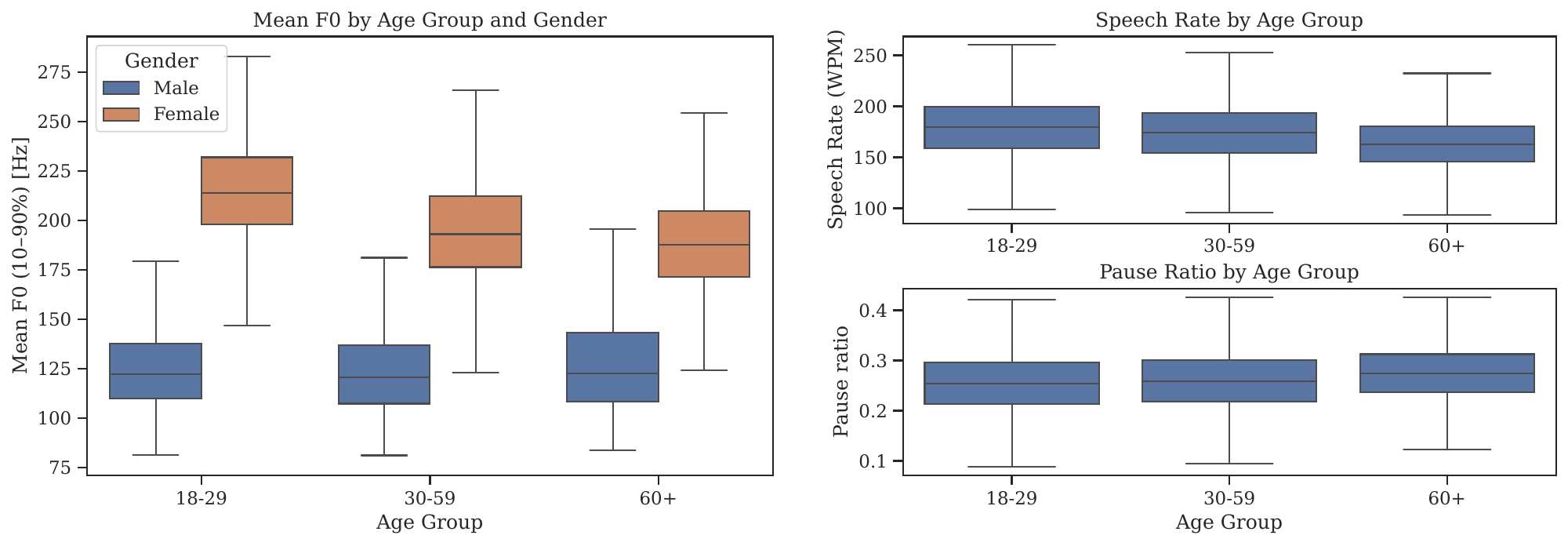}
  \caption{Model-predicted age-bin effects on prosody and rhythm. Left: mean $F_0$ (10--90\% trimmed) across Vox-Profile age bins (18--29 / 30--59 / 60+) stratified by model-predicted sex label ($N{=}118{,}033$). Right: speech rate and pause ratio distributions across the same model-predicted age bins ($N{=}88{,}287$).}
  \label{fig:age_effects}
\end{figure*}

We characterize conversational rhythm with speech rate and pause ratio, capturing how time is allocated between verbal output and silence.
Table~\ref{tab:sex_label_effects} shows that model-predicted sex-label effects on temporal metrics are negligible for speech rate and mean pause duration, and small for articulation rate and pause ratio.
Figure~\ref{fig:temporal_state} shows stronger state dependence, where speech rate increases with arousal and dominance, while pause ratio decreases. 
This coupled pattern is consistent with classic observations that perceived ``speed of talking'' is driven heavily by the structure and frequency of pauses rather than articulation speed alone, motivating joint descriptions of conversational rhythm \cite{goldman1958speech, goldman1961significance}.
It also agrees with applied-linguistics evidence that conversational speech rates occupy a bounded range while varying by situation and interaction \cite{tauroza_speech_1990, nishizawa2024authenticity}.
For high-arousal rows, pooled references flag 15.76\% of speech-rate samples and 13.18\% of pause-ratio samples, while arousal-matched references reduce these rates to 12.50\% and 10.84\%, respectively (Table~\ref{tab:pooled_matched_flags}).

Model-predicted age bin further shifts temporal regimes in a way that matters operationally for evaluation. 
Figure~\ref{fig:age_effects} shows that model-predicted age bins are associated with changes in speech rate and pause ratio, implying that timing targets should not be treated as universal across this operational stratifier. 
This result complements prior conversation analyses showing that timing patterns can index participation style and interactional dynamics \cite{campbell2008individual}. 
For conversational systems, the practical implication is that timing-based evaluation should assess pace and pausing jointly, and condition on model-predicted age bin when that stratifier is available.  

\section{Reference-Based Evaluation Protocol}\label{sec:evaluation_protocol}

The reference regimes in Section~\ref{sec:regimes} are intended to support a lightweight evaluation procedure for S2S dialogue outputs. 
Given an output waveform from a spoken dialogue system, the protocol is:
\begin{enumerate}
    \item Extract the same prosodic and temporal metrics used in this study: $F_0$ mean, $F_0$ SD, $F_0$ range, speech rate, articulation rate, pause ratio, and mean pause duration.
    \item Select a reference stratum using the available conditioning variables, such as model-predicted sex label for absolute $F_0$, arousal or dominance bins for $F_0$ expressivity and rhythm, and model-predicted age bin when timing comparisons require it. If a conditioning variable is unavailable, use the coarsest applicable reference and report that comparison scope explicitly.
    \item Convert each system metric $m$ to a percentile $p_m$ under the selected human reference distribution.
    \item Flag metrics below the 5th percentile or above the 95th percentile as out-of-regime, and report the output as a vector of percentile deviations rather than a single opaque score.
\end{enumerate}
This report identifies which speech-native dimensions are atypical, the direction of the deviation, and the matched conversational regime under which the deviation was measured. 
The protocol is therefore a behavioral plausibility check: it can flag prosodic compression, unusually fast or slow pacing, or atypical pause allocation, but it does not claim to replace human judgments of naturalness or interaction quality.

\section{Limitations}\label{sec:limitations}
This framework provides behavioral plausibility checks rather than a validated perceptual naturalness model. 
The reference regimes and conditioning effects do not yet map deviations to perceptual thresholds or user-rated interaction quality. 
Results are derived from a single English dyadic corpus. 
Operating regions may shift with language, domain, recording conditions, and interaction setting. 
Model-predicted age bin, arousal, dominance, and sex label are derived from Vox-Profile rather than ground-truth speaker-trait metadata, so observed stratification effects should be interpreted as model-conditioned annotations and validated against ground-truth metadata where available.
The analysis uses a binary sex label predicted from voice and hence does not model non-binary or self-identified identity categories.
Future work should validate these regimes against human judgments and system outputs across multiple datasets and languages.

\section*{Acknowledgment}

The preferred spelling of the word ``acknowledgment'' in America is without 
an ``e'' after the ``g''. Avoid the stilted expression ``one of us (R. B. 
G.) thanks $\ldots$''. Instead, try ``R. B. G. thanks$\ldots$''. Put sponsor 
acknowledgments in the unnumbered footnote on the first page.

\bibliographystyle{IEEEtran}
\bibliography{newbib}

\end{document}